\documentclass[oneside,onecolumn]{article}
\usepackage{microtype}

\usepackage[english]{babel}

\def\artid{0001}%

\let\svthefootnote\thefootnote
\newcommand\freefootnote[1]{%
  \let\thefootnote\relax%
  \footnotetext{#1}%
  \let\thefootnote\svthefootnote%
}

\usepackage{amsthm}
\usepackage{thmtools}
\usepackage{thm-restate}

\usepackage{amsmath}

\usepackage[disable,textsize=tiny]{todonotes}
\usepackage[textsize=tiny]{todonotes}

\usepackage{multirow}
\usepackage{booktabs}

\usepackage{tikz}
\usetikzlibrary{arrows.meta}
\usepackage{tabularray}

\usepackage{soul}
\usepackage[utf8]{inputenc} 
\usepackage[T1]{fontenc}
\usepackage{url}
\usepackage{ifthen}
\usepackage{cite}
\usepackage{nicefrac}

\usepackage[hmarginratio=1:1,top=32mm,columnsep=20pt]{geometry} 

\usepackage{balance}
\usepackage{hyperref}
\usepackage{url}

\providecommand{\keywords}[1]{\textbf{\textit{Index terms---}} #1}

\begin{document}

\title{From Abstraction to Reality: DARPA’s Vision for Robust Sim-to-Real Autonomy}

\author{Erfaun Noorani, Zachary Serlin, Ben Price and Alvaro Velasquez}

\date{}
\maketitle

\let\thefootnote\relax\footnotetext{Erfaun Noorani, Zachary Serlin, and Ben Price are with MIT Lincoln Laboratory (emails: {\tt \{erfaun.noorani, zachary.serlin, ben.price\}@ll.mit.edu}). Alvaro Velasquez is with Defense Advanced Research Projects Agency (DARPA) (email: \tt \{alvaro.velasquez\}@darpa.mil). \\

DISTRIBUTION STATEMENT A. Approved for public release. Distribution is unlimited. This material is based upon work supported by the Under Secretary of Defense for Research and Engineering under Air Force Contract No. FA8702-15-D-0001. Any opinions, findings, conclusions or recommendations expressed in this material are those of the author(s) and do not necessarily reflect the views of the Under Secretary of Defense for Research and Engineering.\\

This work has been submitted for publication in AI Magazine.}

\abstract{The DARPA T\textsc{ransfer from} I\textsc{mprecise and} A\textsc{bstract} M\textsc{odels to} A\textsc{utonomous} T\textsc{echnologies} (TIAMAT) program aims to address rapid and robust transfer of autonomy technologies across dynamic and complex environments, goals, and platforms. Existing methods for simulation-to-reality (sim-to-real) transfer often rely on high-fidelity simulations and struggle with broad adaptation, particularly in time-sensitive scenarios. Although many approaches have shown incredible performance at specific tasks, most techniques fall short when posed with unforeseen, complex, and dynamic real-world scenarios due to the inherent limitations of simulation. In contrast to current research that aims to bridge the gap between simulation environments and the real world through increasingly sophisticated simulations and a combination of methods typically assuming a small sim-to-real gap\textemdash such as domain randomization, domain adaptation, imitation learning, meta-learning, policy distillation, and dynamic optimization\textemdash TIAMAT takes a different approach by instead emphasizing transfer and adaptation of the autonomy stack directly to real-world environments by utilizing a breadth of low(er)-fidelity simulations to create broadly effective sim-to-real transfers. By abstractly learning from multiple simulation environments in reference to their shared semantics, TIAMAT's approaches aim to achieve abstract-to-real transfer for effective and rapid real-world adaptation. Furthermore, this program endeavors to improve the overall autonomy pipeline by addressing the inherent challenges in translating simulated behaviors into effective real-world performance.
}


\maketitle

\section{Introduction and Motivation}

Humans possess adaptive expertise—the ability to quickly apply knowledge, plans, and strategies learned in simpler contexts to more complex and unfamiliar situations. An example of adaptive expertise is a firefighter responding to an emergency situation. Although firefighters are trained in various standardized procedures and protocols, real-world scenarios often present unpredictable challenges, such as unusual weather conditions, changing fire patterns, or obstructions in the environment. An experienced firefighter draws on their past knowledge and skills, adapting their approach based on the specific situation at hand. For example, if a firefighter arrives at a fire that has spread rapidly due to high winds, they might adjust their strategy—such as prioritizing certain areas for control or choosing a different approach to ventilation—based on the conditions they face, even if those conditions weren't part of their standard training. The fundamental goal of the TIAMAT program is to enable autonomous systems to develop similar adaptive expertise—quickly creating and executing effective plans based on minimal prior knowledge and generalizing strategies from previous tasks, possibly learned in less complex and more controlled environments, in unpredictable environments with minimal adaptation time. This capability is crucial for deploying autonomous systems in dynamic, real-world environments where they must handle complex, unforeseen challenges.

\textbf{Understanding the Simulation to Reality (sim-to-real) Gap:} 
One foundational hurdle in advancing fielded autonomous systems is translating the success achieved in simulations into practical, real-world situations. The sim-to-real gap refers to the difference between how autonomous systems perform in simulations compared to the real world. While simulations are valuable for training, they cannot fully replicate the dynamic, complex and unpredictable nature of real-world environments. This gap arises because factors such as environmental variability, sensory noise, and complex dynamics are difficult to simulate accurately. While there have been some initial attempts to quantify these gaps, such as through the sim-to-real correlation coefficient \cite{Kadian2020}, they are still mostly notional.

To understand how autonomous systems navigate this challenge, we need to examine the key components that these systems rely on: perception, planning, and control. Perception is how the system senses and interprets its surroundings, such as identifying objects or detecting obstacles. Planning is the process through which the system determines its course of action, like selecting the optimal route or strategy. Control is how the system executes those actions, such as steering, accelerating, or braking. The sim-to-real gap affects each of these dimensions: in perception, simulated sensors may not capture the full range of real-world conditions, such as lighting or weather; in planning, virtual scenarios are often more predictable, leading to overconfidence in decision-making; and in control, real-world dynamics such as tire friction or road surface variations can introduce unexpected behaviors. Together, these gaps make it difficult for systems to seamlessly transition from simulation to real-world deployment.

In our view, bridging sim-to-real gaps can be broken down into the above three types of transfer (perception, planning, and control).
\begin{itemize}
    \item Perception Transfer: This refers to a system's ability to interpret sensory data and semantics in real-world conditions compared to simulations. For instance, a robot trained in a virtual environment might struggle to recognize objects under different lighting, weather conditions, or when those objects are partially damaged, broken, or hidden. A gap in perception transfer could manifest as the robot misidentifying a road sign due to variations in color or shape.
    \item Planning Transfer: This involves a system's capacity to apply strategies to navigate and make decisions in real-world scenarios. For example, a simulated robot may effectively plan a route in an ideal environment but struggle to adapt when faced with unexpected obstacles or noise from a perception system. A gap in planning transfer might be evident if the robot has difficulty rerouting when encountering a previously unknown barrier or must replan when a route is deemed impassable.
    \item Control Transfer: This concerns how effectively a system executes commands in real-world conditions. For example, a quadruped robot that runs smoothly in simulation might struggle with stability in muddy conditions. A gap in control transfer could become apparent if the drone behaves erratically or fails to maintain its intended path when faced with real-world turbulence.
\end{itemize}
Even though we have broken down the transfer into these three components, it is important to recognize that the autonomy stack functions as a system of systems. Its overall performance and transferability depend not only on how well each individual component adapts but also on how effectively these components interact across the entire stack. The integration and seamless functioning of perception, planning, and control are crucial for ensuring that the system operates efficiently in real-world scenarios. Therefore, addressing the sim-to-real gap requires a holistic approach that considers the inter-dependencies among all elements of the autonomy stack.

\textbf{The Limitations of Better Simulations and existing methods:}
Efforts to develop more sophisticated and realistic simulations have undoubtedly contributed to advancements in autonomous systems. Yet, even the most advanced simulations fall short in replicating the full spectrum of real-world conditions. Additionally, existing algorithms, such as domain randomization \cite{Tobin2017}, domain adaptation \cite{You2019}, imitation learning \cite{Hussein2017}, meta-learning \cite{Yu2020}, policy distillation \cite{Czarnecki2019}, and dynamic optimization, typically assume a small sim-to-real gap—an assumption that is not supported, even by the most advanced existing simulations \cite{Zhao2020}.

To improve the transfer of learned autonomy from simulation to reality, two main approaches are commonly pursued: enhancing simulator fidelity and developing algorithms that can effectively learn from low-fidelity simulations \cite{Hofer2021}. However, a major limitation of the high-fidelity simulator approach is the immense amount of time required to learn the necessary autonomy, which is impractical for time-sensitive applications like the 24-hour Air Force planning window \cite{Conner2005}. This gap between simulation and reality underscores a crucial point: merely improving simulation fidelity is insufficient. We, as a community, must also focus on developing algorithms that are data-efficient, robust, and capable of adapting to the disparities between simulated and real-world environments.

\begin{figure}[ht] \label{fig:goal}
\caption{TIAMAT's goal: Increased performance with large transfer gaps}
\centering
\includegraphics[width=0.5\textwidth]{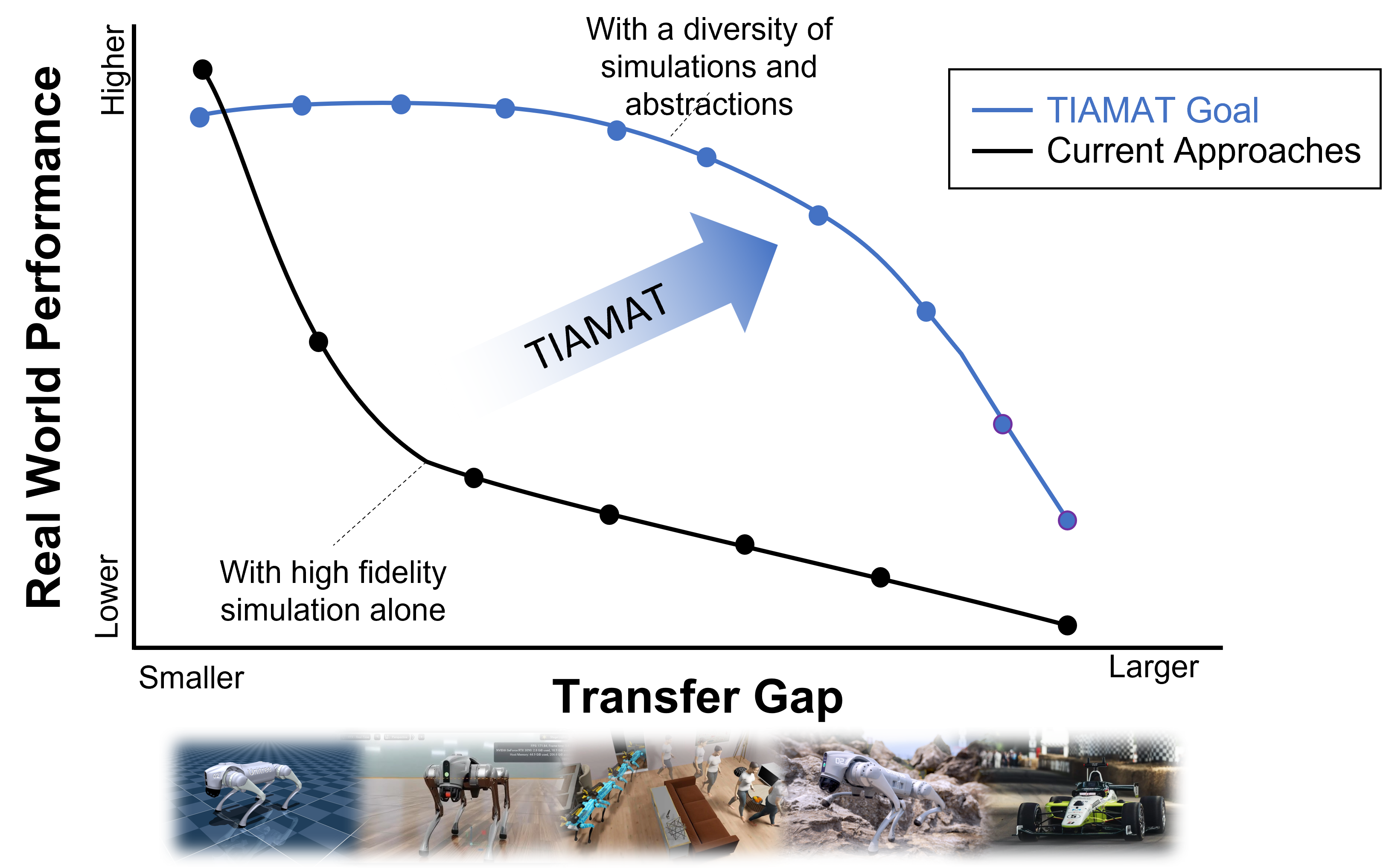}
\end{figure}

\section{ The TIAMAT Program's Vision \& Goals:} 
%
TIAMAT aims to achieve rapid and robust autonomy transfer techniques to enable generalization to environments for which we do not have good models and simulations. Rapid in this context refers to the ability of autonomous systems to operate effectively in real-world environments with limited time to adapt to new dynamics, semantics, context, and environments. As the transfer gap widens between simulated and real-world environments, existing algorithms often experience significant performance degradation. In contrast, the TIAMAT goal is to push out this performance curve as the gap widens, ensuring consistent operational capability despite operational differences. The pictorial representation of this goal is illustrated in Figure \ref{fig:goal}. This approach also aims to minimize the time between development and deployment, accelerating the integration of autonomous technologies into everyday use, no matter the mission of the day.

\subsection{Technical Hypothesis:}
Taking inspiration from foundational models and large language models in the machine learning community, TIAMAT hypothesizes that broader training diversity and experience will lead to more flexible and robust autonomous systems. The TIAMAT program's technical hypothesis centers on the idea that learning from a variety of low-fidelity simulations, in conjunction with limited high-fidelity simulations with shared semantics (e.g., knowledge graphs), leads to rapid transfer from simulation to reality. This hypothesis is informed by the fact that visually and physically different environments can nevertheless share commonalities when viewed through the right level of abstraction. We argue that leveraging a diverse set of low-fidelity simulations, characterized by their inherent imprecision and shared semantics, can foster better generalization. In contrast, the conventional use of a high-fidelity simulation often faces challenges like overfitting (e.g., to simulated physics) \cite{Truong2023} and data drift. By embracing the imperfections of low-fidelity simulations alongside high-fidelity approaches, the program aims to enhance the generalization capabilities of autonomous systems, making them more resilient to real-world variations.

A key insight driving the TIAMAT program is the recognition that diverse environments share underlying semantics. This assumption paves the way for leveraging semantic anchors — abstractions that capture shared semantics across different environments. Semantic anchors offer a common understanding across different environments. For example, various door designs, despite their differing appearances, all fulfill the fundamental function of providing access. These anchors encapsulate key concepts—such as principles of physics or fundamental operational rules—that remain consistent across diverse settings. 

Semantic anchors may be defined by the end user \cite{Velasquez2021}, inferred from the agent's observations \cite{Hasanbeig2021}, or generated based on a learned model of the agent \cite{Carr2021}. Examples of semantic anchors in the literature include logic, natural language, and traffic laws, which play a role in solving problems related to object manipulation \cite{Innes2020}, vision-and-language navigation \cite{Krantz2022}, and autonomous driving \cite{Kiran2021}. In recent studies on semantically-anchored transfer, innovations such as 3D scene graphs have been used for sim-to-real transfer, allowing systems to adapt to previously unseen real-world environments \cite{Velasquez2023b}. Additionally, deterministic finite automata have facilitated sim-to-sim transfer, enabling effective transitions between different grid-world environments \cite{Velasquez2023a}.

By grounding representations in these semantic anchors, the program aims to create robust and transferable autonomous systems that can perform effectively even in the face of significant sim-to-real gaps. For example, the new approach to autonomy transfer using semantic anchors will significantly reduce the data required for learning and transfer by simplifying complex autonomy tasks to more manageable reference points within anchored representations. While large models can have billions of parameters, semantic structures, such as those in logic formulas or language syntax, are much smaller. The effectiveness of semantic structures for efficient transfer learning is demonstrated by large language models, which leverage shared syntax and semantics to transfer knowledge across tasks rapidly through methods like few-shot prompting, chain-of-thought reasoning, and in-context learning \cite{Liu2023,Wei2022,Liu2022}. However, applying this to autonomy presents additional challenges due to variations in platforms, behaviors, and missions across different simulation environments.

\subsection{Technical Challenges}

The primary challenges posed by the TIAMAT program include:

\begin{itemize}
    \item Conducting autonomy transfer under restricted time budgets: Ensuring that autonomy transfer is robust to sim-to-sim and sim-to-real gaps in agent observations, actions, transitions, and goals when there is limited time to adapt to unseen situations and environments.
    \item Refining models and simulations based on agent experience: Establishing a feedback loop between autonomy transfer and abstraction refinement to enhance transfer robustness and effectiveness.
\end{itemize}
These challenges necessitate innovative approaches to decision processes and the development of new methods for sim-to-real transfer. The program encourages the use of multiple methods, including neuro-symbolic transfer using semantic anchors, robust reinforcement learning, physics-informed neural networks, and others.

\section{Abstract-to-Real Transfer}
\emph{Abstract-to-real transfer} is the process by which an agent leverages knowledge, concepts, or models to perform tasks in a real-world environment without explicitly testing in the real-world a priori; specifically where the sim-to-real gap is relatively large. The knowledge gaining process can be accomplished from a wide range of techniques (with sim-to-real transfer being a subset of this process). A difference in abstraction and real-world deployment is expected. However, assuming a broad enough abstraction, the system should be able to rapidly adapt to new unseen environments, platforms, and tasks.

In the process of translating theories and models into practical applications, \emph{refinement} plays a crucial role. Unlike abstract-to-real transfer, which moves knowledge from abstract simulations to real-world environments, refinement involves taking insights and data gained from real-world experiences and applying them back into abstract models or simulations. This iterative process allows for the enhancement of these abstract models, making them more accurate and effective by integrating practical experience. Through refinement, systems can be continuously improved, ensuring that theoretical models evolve in response to actual performance and real-world conditions. This bidirectional approach helps bridge the gap between theoretical predictions and real-world execution, leading to more reliable and adaptable solutions (see Figure 2). 

The TIAMAT program looks to explore and quantify the abstract-to-real transfer gap (i.e., how reasoning about complex concepts transfers to physical system actions) with the goal of enabling general understanding of abstract concepts in real world autonomous systems. Semantics in particular are fundamental to understanding the physical world around us with different contexts and physical embodiments dictating how they are interpreted. The semantic abstract-to-real gap is caused by differences between a concept (e.g., object labels, places, ideas) and their more complex physical embodiment. Practically, concepts can be expressed in many abstractions (e.g., in a knowledge graph or large language model). 


\begin{figure}[h] \label{fig:a2r_loop}
\caption{Abstract to Real Transfer Cycle}
\centering
\includegraphics[width=0.5\textwidth]{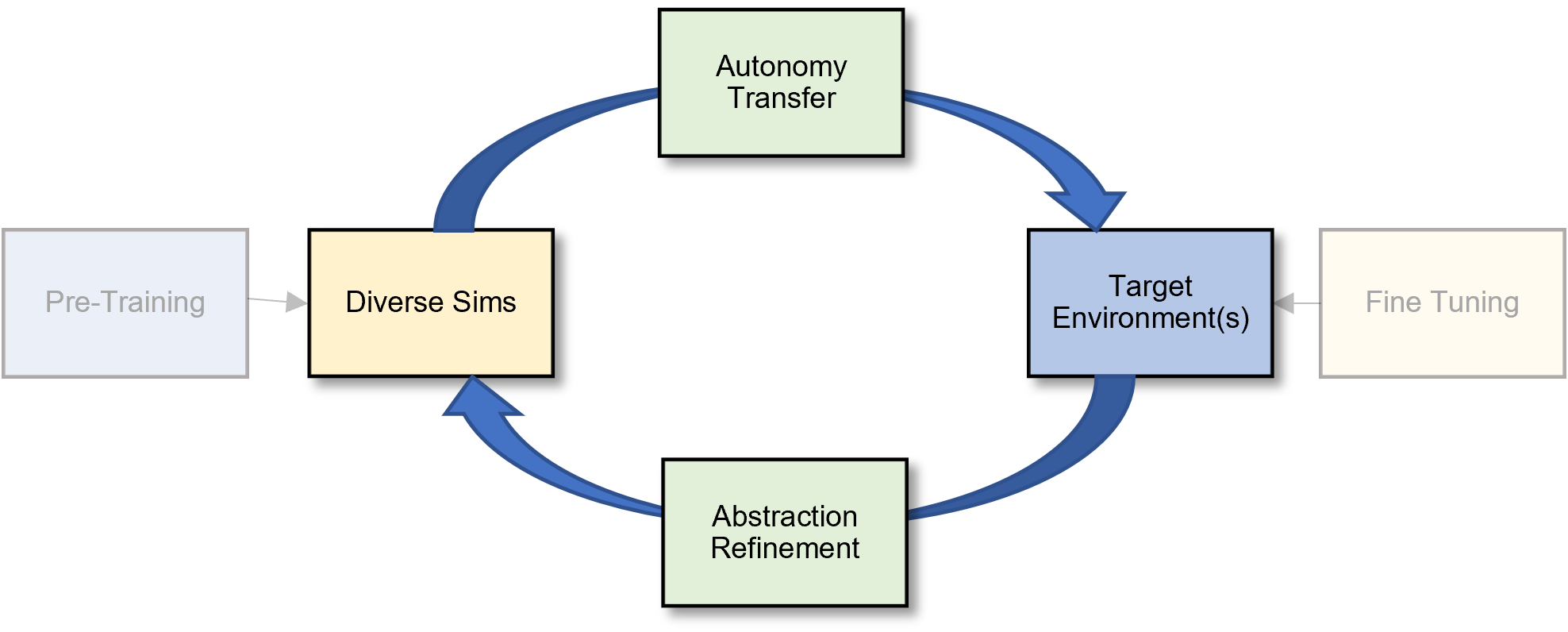}
\end{figure}

\section{Current Approaches to Rapid and Robust Autonomy}
Rapid and robust autonomy is an emerging priority across various industries, emphasizing the need for systems that can achieve full autonomous operation on novel tasks within a single day or faster. To realize this goal, a combination of methodologies are being employed, including symbolic approaches, data-driven strategies (e.g. machine learning), and, increasingly, generative methods such as large language models (LLMs) and vision-language models (VLMs). Most modern systems integrate elements from these diverse approaches, creating robust solutions through the cross-fertilization of ideas.

Symbolic approaches rely on rule-based reasoning, enabling systems to make decisions based on established criteria. In contrast, data-driven methods typically leverage machine learning to adapt and improve performance by analyzing large datasets and approximating a latent embedding for decision making. Generative methods, including LLMs and VLMs, play a crucial role by creating simulations and predictive models that enhance decision-making processes. They leverage extremely large datasets to approximate bulk knowledge. Meanwhile, neuro-symbolic approaches combine symbolic reasoning with deep learning, allowing for a more nuanced understanding of complex scenarios and potentially more predictable systems.

These methodologies operate across all parts of the autonomy stack—from perception to planning and control. For example, in the perception layer, large models like Contrastive Language-Image Pre-training (CLIP) analyze visual inputs to identify and interpret objects using an open vocabulary in real time. For planning, symbolic logic can formulate strategic decisions based on this interpreted data. Adaptive and robust reinforcement learning (RL) techniques are employed in the control layer, ensuring that autonomous systems can effectively respond to dynamic environments. This combination of strategies enhances the overall effectiveness of robust same-day autonomy, enabling  solutions that meet the demands of rapidly changing operational contexts.

\section{TIAMAT Program Challenge Competitions:}

The TIAMAT program is creating two competitions to explore aspects of the abstract-to-real problem and evaluate performers. The Abstract Planning and Semantic Understanding (APSU) Challenge emphasizes navigating and interacting within semantically rich environments, focusing on how well the robot can understand and act within complex real-world settings that were previously unseen by the system. The Indy Autonomous Challenge concentrates on precise control and planning transfer of autonomous Formula One vehicles, evaluating how quickly and efficiently the autonomous system can adapt in high-speed racing scenarios where rapid and precise adaptation is key.

In line with abstract-to-real transfer, closing the loop and refining solutions and abstractions will be important for prototypes to excel as real world capabilities. Ideally, being able to refine semantic, dynamic, and mission understanding will allow the prototype solutions to adapt more effectively to new experiences. To this end, the TIAMAT challenge competitions will allow teams time to refine their solutions within a fixed pre-mission and pre-scoring period on the day of the challenge. The spirit of this adaptation period is not to have multiple solutions that are switched between, but instead have a prototype that is flexible and can refine its performance through limited experience.

 The developed systems will be tested with transfers between multiple environments and tasks while maintaining a common set of semantic anchors. The required adaptation will be characterized by three broad categories: (1) Perception and Semantics, (2) Dynamics and Terrain, and (3) Missions and Tasks. The perception and semantics adaptation will necessitate that the prototype can transfer between primarily environmental semantics (e.g., training in an indoor or desert environment and deploying in a lush forest). The dynamics and terrain adaptation will test the prototype's ability to navigate various terrains and physical interactions (e.g., grass, sand, mud, and concrete with a legged robot) or adapt to unexpected system dynamics (e.g., friction coefficients between a tire and a racetrack). Lastly, the mission and task adaptation will test the reasoning capabilities of the system by giving it a high level description of various missions that require the agent to extract important details and act on those extracted details (e.g., the agent could be asked to search a disaster scene for survivors and report back the location of all the injured humans wearing blue shirts). Each of these adaptation categories is coupled with the others, but taken as a whole they necessitate a flexible prototype with broad knowledge and rapid adaptation to new environments and tasks.

\begin{figure*}[h]
\caption{APSU Challenge Timeline}
\centering
\includegraphics[width=0.98\textwidth]{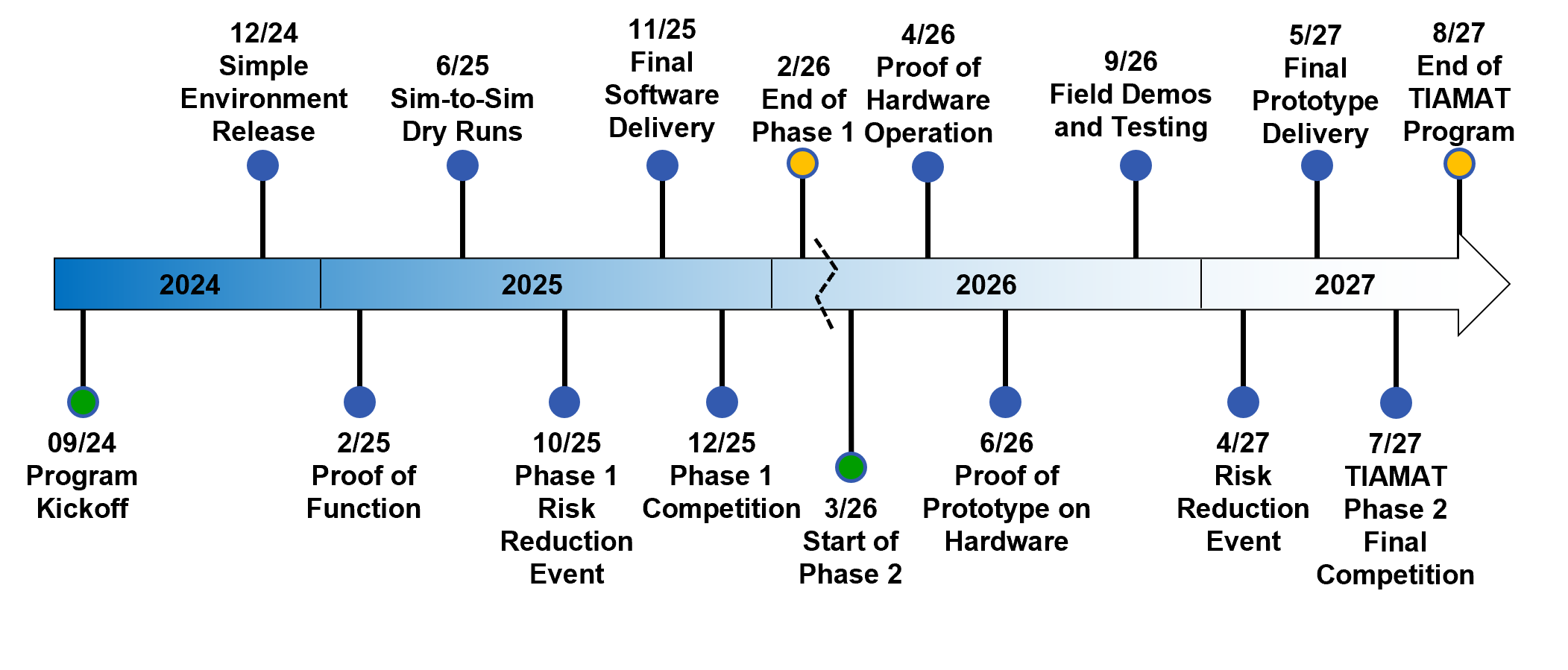}
\end{figure*}

\subsection{Abstract Planning and Semantic Understanding (APSU) Challenge:}

This challenge competition focuses on semantic understanding and transitioning agents from simple abstractions and indoor simulation environments, to outdoor simulations and eventually hardware that must adapt to changes in scene appearance, variability in terrain and control, and potentially reason about how current planning actions will impact future outcomes in semantic richness and terrain traversability. This challenge is designed to foster innovations that enable quadruped robots to seamlessly transition from simulated environments to real-world applications with minimal adaptation time.

The first phase of this challenge will test prototypes in simulation only. Performers are given a variety of basic simulators: MiniGrid, Mujoco, Habitat, and basic IsaacLab instances. The Phase 1 challenge will give performers exemplar scenarios in advance of the final evaluation. At the Phase 1 evaluation event, performers will be given unseen indoor and outdoor environments and an hour to explore the environment and hone their system's understanding. The performers will also be given a sample prompt for that new environment. At the end of that hour, the system will be given a new prompt for both the indoor and outdoor scenarios and be graded for that run.

The second phase of this challenge will focus on real-world prototype performance. The competition plans to use a Unitree Go2 Quadruped robot. The final challenge will take place at an undisclosed location with teams getting 8 hours of practice time in the terrain to fine-tune their hardware systems. The final challenge will have a corresponding simulation environment and metrics will evaluate the performance difference between abstractions and the real-world systems. See Figure 3 for more detail on program milestones and timeline. 

\subsection{Indy Autonomous Challenge}

This challenge will feature an autonomous racing component inspired by Formula 1, where cutting-edge technologies and techniques in autonomous driving will be tested and refined. This high-speed racing environment will provide insights and real-time data to advance our understanding and implementation of autonomous systems in dynamic and high-stakes scenarios.

\section{Evaluation \& Metrics}
Metrics will be used to evaluate the APSU challenge prototypes. As Phase 1 progresses, more detailed and concrete metrics will be developed for evaluating the Phase 1 and 2 Challenge prototypes. Below is a (non-exhaustive) list of likely metrics for evaluation of the Phase 1 (Sim-to-Sim) challenge prototypes.

\begin{itemize}
    \item \textbf{Semantic Similarity:} The semantic similarity metric indicates the semantic distance between two objects (regions, concepts, or environments) in a common concept embedding space. A potential example of this metric could include cosine similarity of CLIP embedding features \cite{edge2024local}. This metric will be used to assess how closely the objects found in the APSU challenge matched the prompt and semantic anchors provided (e.g. how closely a mug \textit{response} matches a wineglass \textit{prompt}). 
    \item \textbf{Runtime:} Total wall clock execution time of a submission to the challenge given a prompt, semantic anchors, simulation instance, and solution prototype. This is for a single instance of the Phase 1 challenge and is external from the adaptation time.
    \item \textbf{Adaptation Time:} The total wall clock time used by a challenge team to fine-tune their prototype to perform in the challenge environments. 
    \item \textbf{Inference Time:} The time a prototype solution takes to compute a single output action given a single input observation. Performers may leverage inference-time computation if they wish to as is commonly done with so-called large reasoning models.
    \item \textbf{Semantic Accuracy:} Given an expected output list of items to be found during the challenge, this metric evaluates how many of those objects match the correct answer (i.e., the difference between the expected output objects list and returned solution objects list) using a semantic similarity metric.
    \item \textbf{Semantic Localization:} Given an expected output list of items and their general semantic locations (e.g., wineglass in the dinning room), this metric evaluates how accurately the challenge prototype inferred abstract locations within the environment and communicated that in the solution. 
    \item \textbf{Localization Accuracy:} Given an expected output list of items and their location within the test environment, this metric measures how far from the desired object's location the prototype reported the objects to be. 
\end{itemize}

\section{Performers Approaches:}
\begin{figure*}[h]
\caption{Performer approaches in the autonomy transfer and refinement loop}
\centering
\includegraphics[width=0.98\textwidth]{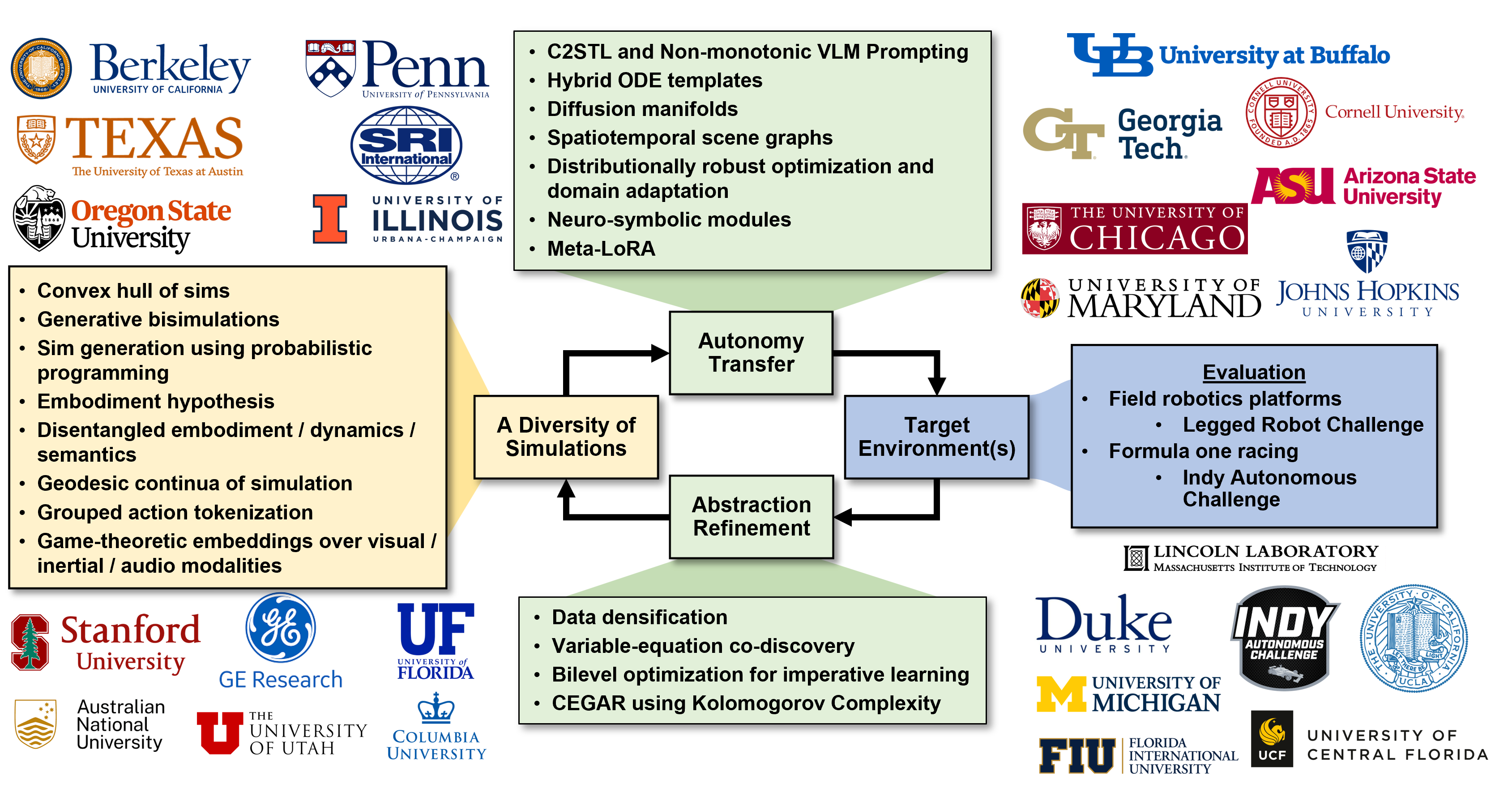}
\end{figure*}

Participants in the program are expected to demonstrate proficiency in transferring their semantic understanding, as well as their planning, and control capabilities from simulated environments to real-world applications.

The performers look to explore two complementary directions aimed at advancing the autonomy and adaptability of autonomous systems. First, autonomy transfer methods focus on enhancing the ability of systems to autonomously learn, adapt, and transfer knowledge across different tasks and environments. These approaches seek to improve how systems generalize their learning, enabling them to perform effectively in new, unseen contexts. Second, abstraction refinement methods emphasize the importance of refining and adapting the internal representations of models for more efficient learning and problem solving. Performers may focus on one or both of these directions, depending on their expertise, with some specializing in autonomy transfer methods and others in abstraction refinement methods, or even combining both to enhance system adaptability and knowledge transfer.

Beyond these core approaches, several advanced methods have been proposed by performers that aim to further refine both the forward and reverse directions. For example, convex hull simulations and generative bi-simulations offer new ways of simulating and testing models in complex environments, helping to evaluate the robustness of learned behaviors. Simulation generation using probabilistic programming provides a more flexible way to model uncertain environments, improving the ability of systems to adapt to variability. Concepts like the embodiment hypothesis and disentangled embodiment focus on how physical interaction and sensorimotor experience can be used to improve learning, while geodesic continua of simulation provide a mathematical framework for better understanding the dynamics of complex systems. Grouped action tokenization and game-theoretic embedding across multiple modalities—such as visual, inertial, and audio data—open new possibilities for creating multi-modal systems that can better replicate real-world interactions. 

\subsection{Performers}
TIAMAT has 14 performers, each led by a different university or industry partner. The lead institutions include (in no specific order) Florida International University, University at Buffalo, Oregon State University, University of Pennsylvania, University of California at Berkeley, University of Central Florida, University of Maryland, Johns Hopkins University, GE Research, University of Michigan, SRI International, University of Texas at Austin, and Duke University. 

The performer approaches can be grouped broadly across the three transfer and adaptation categories of perception and semantics, missions and tasks, and dynamics and terrain. Most of the proposals addressed all three categories, though some teams have a stronger focus in one or two areas. For examples on \textit{Perception and Semantics}, teams from Florida International University and University at Buffalo focus on enabling the system to transfer across different environmental semantics, such as training in one domain and deploying in another with different visual and linguistic representations. University of Maryland and Johns Hopkins University also propose leveraging scene graphs and symbolic reasoning to bridge perceptual gaps. In the \textit{Missions and Tasks} category, University of California Berkeley, SRI International, and Duke University tackle the challenge of transferring knowledge between tasks, using approaches such as reinforcement learning, game-theoretic models, and task-specific decoupling mechanisms. For examples in \textit{Dynamics and Terrain}, Oregon State University and University of Michigan address the adaptation required for diverse terrains and physical interactions by incorporating dynamic models and uncertainty quantification into their approaches. Across the categories, teams such as GE Research provide comprehensive frameworks that span all three with equal emphasis, focusing on semantic anchors, domain adaptation, and knowledge transfer for multi-fidelity environments. In the next section, we will explore the different proposed approaches and contributions from each selected team.

\textbf{Florida International University Team} 
\newline
This team is pursuing a neuro-symbolic approach to semantic transfer that leverages the state-of-the-art in diffusion models for video and for autonomy, large multi-modal
models, visual-language models for autonomy, and verbal reinforcement learning, and couple them with new ideas in this context, such as parameterized probabilistic hybrid automata,
equalizers, functors, Galois connections, visual reinforcement learning, as well as Kolmogorov complexity for synthesizing abstract world models. The team addresses abstract world model transitions using probabilistic hybrid automata, while they employ Kolmogorov complexity to understand action sequences. 
They explore probabilistic temporal logics and parameterized probabilistic hybrid automata as semantic anchors. The FIU team also contains participants from the University of Florida.%

\textbf{University at Buffalo, SUNY Team} 
\newline
This team proposes a self-supervised neural-symbolic reasoning framework that capitalizes on the guidance provided by a symbolic reasoning process, grounded in bi-level optimization, in conjunction with common-sense-infused reciprocal learning to design a rapid and transparent transfer paradigm with a unique mechanism for learning with memory. Their approach enables a new knowledge transfer paradigm beyond conventional costly backpropagation that can quickly adapt to new environments based on element-wise operation and explicitly inherit commonsense rules, resulting in rapid and transparent knowledge transfer. The proposal emphasizes observations with causal graphs to link observed events to other observations and proposes a self-supervised imperative learning approach grounded in bi-level optimization. This includes long-term common-sense learning, short-term transfer learning, and real-time adaptation. 

\textbf{Oregon State University Team} This team presents a balanced approach to bisimulation to effectively break down environment transfer into observation and action spaces. The proposal addresses TIAMAT problems through a bi-step transfer process: 1) Transfer Preparation
Stage: learn transfer knowledge about source-target relationships via model interactions, pre-existing data, and prior knowledge, and 2) Transfer Application Stage: use the transfer knowledge under a time bound to solve novel target tasks. The team uses generative bisimulations to model transition dynamics to handle transfer between robotics dynamics models efficiently. Their stagewise approach, with distinct preparation and application phases, aims for rapid and accurate adaptation. The OSU team also contains participants from the University of Utah.%

\textbf{University of Pennsylvania Team} This team focuses on vision-based, high performance, autonomy systems capable of strategic reasoning from abstract models. They plan to do this by integrating game-theoretic embeddings to model actions and goals, aiming to maximize Nash equilibrium. Their method includes game-theoretic inference, multi-modal abstractions, and in-context learning for adaptation. The proposal leverages the team’s substantial experience in drone racing, legged robotics, and autonomous racing. The UPenn team also contains participants from the University of Illinois Urbana-Champaign and the University of California, Berkeley.%

\textbf{University of California Berkeley Team} This team proposes a comprehensive probabilistic programming toolkit for all components. It plans to integrate high-performance tools like Scenic into an ecosystem for transfer learning, addressing both abstract simulations and high-fidelity environments. The team addresses several critical topics, including robust transfer using LLMs, reinforcement learning (RL), and safe control. They propose innovative approaches such as combining RL with control barrier functions (CBFs) for enhanced safety and efficiency in learning. The team’s extensive experience spans various domains, including IAC F1 racing, legged locomotion, manipulation, and UAV control. The Berkeley team also contains participants from Stanford University. %

\textbf{University of Central Florida Team} 
This team looks to combine distributionally robust optimization with domain adaptation to manage changes in transition dynamics. The team leverages a distributionally robust optimization framework and plans to apply it to supervised learning and reinforcement learning. Unlike common approaches (aligning distributions and extracting domain-invariant features), the proposed approach constructs uncertainty sets of distributions around both the source and target domain knowledge. %

\textbf{University of Maryland Team} This team employs meta-action embeddings and scene graphs for observations. The proposal combines domain-transferable robust perception with abstract symbolic reasoning abilities and hierarchical planning. This approach exploits abstract knowledge represented through
scene graphs, knowledge graphs, and physics, in addition to world knowledge from LLMs. It will be made robust at training time through stress-testing and adversarial training and follows the offline-pretraining-online-adaptation framework of modern foundation models, making it applicable to a broad range of domains. The UMD team also has participants from the University of Chicago and University of Texas at Austin.%

\textbf{Johns Hopkins University Team} This team extends a Low-Rank adaptation fine-tuning method and uses spatiotemporal scene graphs for predicting transitions and goals. The team leverages a scene graph generator parsing observations into graphs, a multi-modal Large Language Model (LLM) contextualizing graphs into semantic relationships, a vision language planning model (VLPM) fusing visual and symbolic information for decision making, a planner and controller integration (PCI) module issuing low-level controls, and a meta-learning adapter rapidly tuning LLM-based components. The JHU team also has participants from the Georgia Institute of Technology, Cornell University, and Arizona State University.%

\textbf{GE Research Team} This team seeks to take geometric and topological perspectives to model environments. The team proposes a framework based on foundation models for establishing semantic anchors between diverse simulation environments. Their approach uses similarity measures between simulators to allow for the construction of a domain topology where simulators are situated with respect to a critical set of knowledge axes. Then, meta-learning over a set of low-fidelity environments will result in expertise regarding how best to perform such transfers. Instead of attempting to transfer modular knowledge directly between simulators, the GE team will construct a new kind of knowledge escalator in the form of simulation continua. Physics anchors will be used for the interpolation of state-transition models between simulators and use image manifold methods to enable the interpolation of observation models. These geodesic simulation continua will enable the gradual transfer of expertise between simulators (both low- and high-fidelity) in time periods significantly shorter than the current state-of-the-art permits. The GE team also has participants from the University of Pennsylvania, Australian National University, and University of California Los Angeles. %

\textbf{University of Texas at Austin Team} This team proposes a compositional neuro-symbolic learning approach for sequential decision-making. They propose composing modular encoders that cover a large range of possible domains that are learned a priori and then reused to facilitate a minimal adaptation requirement. At runtime, they propose combining multiple modules to obtain a policy that maximizes transfer. Finally, they propose a framework for refining policies given new domains using model abstraction techniques from both reinforcement learning and formal methods. %

\textbf{University of Michigan} This team decomposes the transfer problem into two parts: a central model with general knowledge and capabilities, and a smaller model targeted at addressing gaps between training and target domains, using Bayesian inference and data densification to refine the models. The proposal centers around the development of the two modules that will enable rapid autonomy transfer: (1) an uncertainty quantified perception module using LLMs that leverages language-based grounding
to identify general and extensible sim-invariant features and (2) a domain gap-filling planning module that can be efficiently trained by identifying domain gaps in sim-to-sim and sim-to-real transfer and apply dense learning method developed from our prior work to overcome domain gaps. The UMich team also has participants from Neya Systems, LLC.   %

\textbf{SRI International} This team proposes to use VLMs and chance-constrained signal temporal logic (C2STL) for observations and transitions. To transfer autonomy robustly, the approach involves: (1) using embodied interactions to develop hierarchical, functionality-based representations as stable conceptual anchors; (2) applying non-monotonic reasoning to handle novel entities and behaviors with logical frameworks that support the contradiction of prior knowledge; (3) utilizing spectral graph analysis to identify useful task-agnostic action sequences from an abstract model; (4) accelerating fine-tuning in the target domain by leveraging abstract strategies to define sub-goals and shape rewards, enhancing reinforcement learning efficiency; and (5) designing a training curriculum and tuning environment parameters based on out-of-distribution detection and domain discrepancies to effectively bridge the sim-to-sim or sim-to-real gaps. %

\textbf{Duke University} This team proposes creating foundation models for observations, actions, and transitions and uses symbolic regression for reverse refinement. The team proposes novel decoupling mechanisms on embodiment versus task-specific and dynamics versus semantics-specific modules. Two mechanisms are tightly connected to learn reusable and fast adaptable modules. To capture the shared fundamental physical principles between environments, the team's method integrates hybrid Ordinary Differential Equations (ODEs) with
neural networks and employ meta-learning strategies to significantly enhance the adaptability, efficiency, and effectiveness of robotic systems. 3) They also propose leveraging shared semantic knowledge across different environments and propose a novel algorithmic framework to decouple common-sense semantic knowledge with actionable primitives that are typically embodiment specific.
\newline

In summary, the proposals present a diverse array of approaches to addressing the challenges outlined in this program. From theoretical explorations of robust reinforcement learning and transfer to practical implementations in F1 racing, drones, and quadrupeds, each proposal offers unique strengths.

\section{Conclusions/Discussion}
The TIAMAT program represents a bold step forward in the development of autonomous systems capable of rapid and robust transfer across diverse environments and missions. By leveraging the strengths of low-fidelity simulations and semantic anchors, the program aims to overcome the limitations of traditional high-fidelity approaches, paving the way for more adaptable and resilient autonomy in dynamic real-world scenarios.

\vspace*{-12pt}

\section*{Acknowledgments}
This work was funded by the Defense Advanced Research Projects Agency under the Transfer from Imprecise and Abstract Models to
Autonomous Technologies (TIAMAT) program.

\bibliographystyle{unsrt}
\bibliography{refs} 
\nocite{*}
\end{document}